# Two-Stream 3D Convolutional Neural Network for Human Skeleton-Based Action Recognition

Hong Liu, *Member, IEEE,* Juanhui Tu, *Student Member, IEEE,* Mengyuan Liu, *Student Member, IEEE,*

*Abstract*—It remains a challenge to efficiently extract spatialtemporal information from skeleton sequences for 3D human action recognition. Although most recent action recognition methods are based on Recurrent Neural Networks which present outstanding performance, one of the shortcomings of these methods is the tendency to overemphasize the temporal information. Since 3D convolutional neural network(3D CNN) is a powerful tool to simultaneously learn features from both spatial and temporal dimensions through capturing the correlations between three-dimensional signals, this paper proposes a novel two-stream model using 3D CNN. To our best knowledge, this is the first application of 3D CNN in skeleton-based action recognition. Our method consists of three stages. First, skeleton joints are mapped into a 3D coordinate space and then encoding the spatial and temporal information, respectively. Second, 3D CNN models are seperately adopted to extract deep features from two streams. Third, to enhance the ability of deep features to capture global relationships, we extend every stream into multitemporal version. Extensive experiments on the SmartHome dataset and the large-scale NTU RGB-D dataset demonstrate that our method outperforms most of RNN-based methods, which verify the complementary property between spatial and temporal information and the robustness to noise.

## I. INTRODUCTION

HUMAN action recognition has been widely applied in various applications, including intelligent surveillance [1], human-computer interaction [2], and video analysis [3]. The 3-D representations of human actions provide more comprehensive and discriminative information than 2-D RGB videos. During recent years, the skeleon-based 3D action recognition has been attracting increasing attention due to its high level representation and robustness to appearances and surrounding distractions [4], [5], [6], [7].

Recently, Recurrent Neural Networks (RNN) models and Long-Short Term Memory (LSTM) neurons [8], [9] have been used to model temporal evolutions of skeleton sequences [10]. These RNN-based methods tend to overstress the temporal information [11]. However, the most effective action recognition needs the combination of spatial and temporal information. Considering that 3D convolutional neural network can extract correlations between high-dimensional signals by performing 3D convolutions [12], it has been established as a natural and suitable choice for action recognition, object recognition[13], vehicle detection [14] and human pose estimation [15] to receive a 3-dimensional input. Therefore, this

The authors are with the Key Laboratory of Machine Perception(Ministry of Education), Shenzhen Graduate School, Peking University, Beijing, China(e-mail:hongliu@pku.edu.cn;juanhuitu@sz.pku.edu.cn;mengyuanliu@sz.pku.edu.cn)

paper proposes a novel two-stream 3D CNN model, which intends to reinforce the spatial and temporal information simultaneously.

First, we use a sequence-based transform method proposed by Liu et al. [16], which eliminates the effect of view variations. Further, the transformed skeleton joints for every action sequence are mapped into a 3D coordinate space. Second, to facilitate 3D CNN to learn robust features, skeleton joints are encoded into spatial volume and temporal volume respectively through encoding spatial and temporal information. Third, two-stream 3D CNN separately captures the spatial and temporal information at a fine temporal scale, which enhance the spatial-temporal features. Finally, we convert original skeleton sequences into multi-temporal sequences to capture large scale of temporal information.

## II. RELATED WORK

### A. RNN-based Methods

Most recent action recognition methods are based on Recurrent Neural Networks and Long-Short Term Memory in some form. Du *et al.* [10] proposed an end-to-end hierarchical RNN to encode the relative motion between skeleton joints. Skeletons are split into anatomically-relevant parts, which are fed into each independent subnet to extract local features. Since LSTM can learn long-term and short-term dependencies in the input sequences using special gating schemes, many works chose LSTM to learn features. Shahroudy *et al.* [17] proposed a part-aware LSTM which has part-based memory sub-cells and a new gating mechanism, showing that LSTM outperforms some hand-crafted features and RNN. Zhu *et al.* [18] used an end-to-end fully connected deep LSTM network to learn the co-occurrence features of skeleton joints. However, RNN-based methods tend to focus on the representation of temporal information [11].

### B. 3D CNN-based Methods

3D CNN was proposed for human action recognition [12], [19]. 3D convolutional layer takes a volume as input and outputs a volume. Both spatial information and temporal information are abstracted layer by layer. Tran *et al.* [20] proposed a simple, yet effective approach for spatial-temporal feature learning using 3-dimensional convolutional neural network, which verifies that 3D CNN can achieve faster and accurate performance. Especially, the features used in [20] have four properties for an effective video descriptor: generic, compact, efficient and simple. Cao *et al.* [21] provided an more effective and robust joints-pooled 3D deep convolutional descriptor (JDD), generating promising results on real-world datasets. In general, 3D CNN can automatically capture correlations between three-dimensional signals thereby exploring distinctive spatial-temporal information.

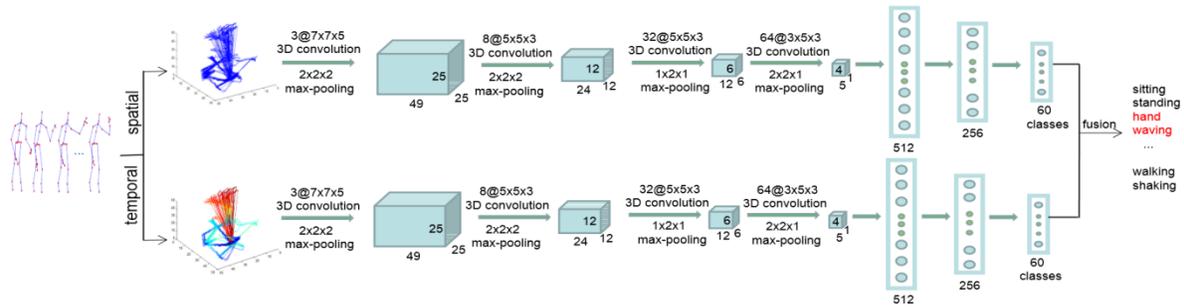

Fig. 1: Overall pipeline of the proposed method. Given an action sequence as input, then encoded into spatial volume and temporal volume respectively after some preprocessing steps. Next we use a two-stream 3D CNN network to learn features. This architecture consists of 4 convolutional layers, 4 max-pooling layers, and 2 fully connected layers. Detailed descriptions are given in the next. Finally the two-stream network are fused by element-wisely multiplying their respective class-membership probabilities.

Our main contribution lies in two aspects: (1) we originally propose a two-stream 3D CNN model, which are mutually compensated and robust to noise. (2) An improved 3D CNN based on [22] which especially can avoid overfitting and has less parameters than C3D [20].

### III. THE PROPOSED METHOD

#### A. Encoding Spatial and Temporal Information

Since different views impact the appearance of skeletons, we adopt a spatial transform proposed by Liu *et al.* [16] as a preprocessing step to solve the problem of viewpoint variations. Assuming an action $H$ has $F$ frames and each skeleton consists of $M$ joints, the *m-th* skeleton joint on the *f-th* frame is formulated as $p_m^f = (x_m^f, y_m^f, z_m^f)$, where $f \in (1,...,F)$, $m \in (1,...,M)$. We use the joint configuration in the NTU RGB+D dataset [17], where $M$ equals to 25. Then, since there are few marked points for an action sequence, we perform the interpolation operation between the consecutive joints to enrich joint information. Next, skeleton joints from an action sequence are mapped into a 3D coordinate space $D$ and then encoded into spatial volume and temporal volume separately. Particularly, it is not only useful for 3D CNN to capture correlations but also solve the problem of inconsistent frames for every skeleton sequence to retain complete information. Let $F_s(x_m^f, y_m^f, z_m^f)$ be spatial value in spatial volume indicating regions of motion, which represents the encoded spatial information. Then, $F_s(x_m^f, y_m^f, z_m^f)$ is defined as:

$$F_s(x_m^f, y_m^f, z_m^f) = \{1, p_m^f \in D; 0, otherwise\} \quad (1)$$

Considering that it is difficult to recognize two actions with similar motion regions but reverse temporal orders, such as actions *"standing up"* and *"sitting down"*. Therefore, let $F_t(x_m^f, y_m^f, z_m^f)$ represents time value in time volume to distinguish them. Its value changes as frame number increases. For the results presented here, we use a simple replacement define as:

$$F_t(x_m^f, y_m^f, z_m^f) = norm(f) \quad (2)$$

where function *norm* indicates that $F_t(x_m^f, y_m^f, z_m^f)$ is normalized to [0,1]. Compared with $F_s(x_m^f, y_m^f, z_m^f)$, $F_t(x_m^f, y_m^f, z_m^f)$ encodes temporal information of action sequences. As shown in Fig. 2, it illustrates the difference between spatial encoding and temporal encoding of action *"hand waving"*. It can be seen that temporal encoding captures the temporal variation. The deeper the color, the more backward the time sequence. Consequently, fusing the spatial and temporal features reinforces each other to achieve better performance.

#### B. Two-Stream 3D CNN Model

We use the network for 3D convnet which is inspired by [22]. As shown in Fig. 1, the architecture of spatial stream is the same as that of temporal stream. For individual stream, the 3D CNN network is comprised of 4 layers of 3D convolution, each followed by a max-pooling, and 2 fully connected layers. The filter numbers for each convolution layer are 3, 8, 32 and 64 respectively and 512, 256 neurons for fully connected layers separately. Same as [22], we use filters of kernal size of 7x7x5, 5x5x3, 5x5x3, 3x5x3 respectively for convolutional layers. Particularly, to reduce overfitting and improve the generalization of classifier, Molchanov *et al.* [22] proposed an effective spatio-temporal data augmentation method to solve the problem. We add dropout layer [23] between the convolution layer and the max-pooling layer to eliminate overfitting. In addition, we use padding after the first three convolution layer in order to make sure the input size is equal to the output size of the convolution operation. The experimental results verify the effectiveness of this improvement. Most importantly, the full model has 910k parameters, which is far less than C3D model [20].

All classification outputs are softmax activated and trained with cross-entropy loss. Furthermore, for activation functions, all the layers in the networks use the rectified linear unit (ReLU):

$$f(z) = max(0, z) \quad (3)$$

We train the spatial stream and temporal stream separately and merge them only during the forward propagation stage employed for decision making. For each network, with parameters $W_S$ and $W_T$ respectively, it produces class-membership probabilities ($P(C|x, W_S)$, $P(C|x, W_T)$) for classes $C$ given the action's observation $x$. We multiply the class-membership probabilities element-wisely from the two-stream network to compute the final class-membership probabilities for the action recognition classier:

$$P(C|x) = P(C|x, W_S) * P(C|x, W_T) \quad (4)$$

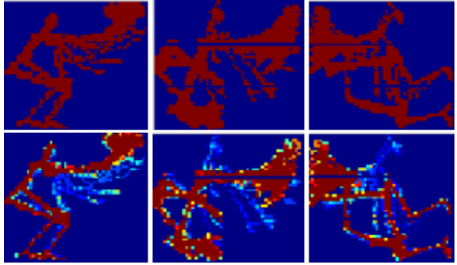

Fig. 2: Comparison between spatial volume (the upper) and temporal volume (the lower) of action *"hand waving"* from three orthogonal planes.

Then, the class label of $c^*$ can be obtained as follows:

$$c^* = argmax \, P(C|x) \quad (5)$$

*C. Multi-Temporal Structure*

We can implement 3D CNN model with different scale of convolutional filters to extract more discriminative information and capture large scale of temporal information. However, this way adds complexity of the 3D CNN model. This paper converts original skeleton sequences into multi-temporal sequences, and then uses two-stream 3D CNN model to extract deep features respectively. As shown in Fig. 3, 3D volume represents the volume of encoding spatial and temporal information. Then multi-temporal 3D volumes are trained by the two-stream 3D CNN model respectively and finally fused to get the result. More exactly speaking, given a skeleton sequence with $F$ frames, "Level 0" represents the entire skeleton sequences; "Level 1" represents the subsequence from the beginning to the $[F/2] - th$ frame; "Level 2" represents the subsequence from $[F/4]-th$ frame to $[3F/4]-th$ frame; "Level 3" represents the subsequence from $[F/2]-th$ frame to the end. 3D volumes extracted from different temporal levels not only capture the multi-scale specific local patterns, but also enhance the global relationships.

## IV. EXPERIMENTS AND ANALYSIS

*A. Datasets*

*1) NTU RGB+D Dataset:* This dataset contains 56880 sequences (with 4 million frames) of 60 classes performed by 40 subjects and captured by three cameras. It is a very challenging dataset due to sequence length, reverse time series action pairs and noisy skeleton joints. Some snaps are shown in Fig. 4. To ensure a fair comparison, we follow the two standard protocols proposed by Shahroudy *et al.* [17]. In crosssubject evaluation, we split the 40 subjects into training and testing groups. Each group contains 20 subjects. For crossview evaluation, we pick all samples of camera 1 for testing and samples of cameras 2 and 3 for training.

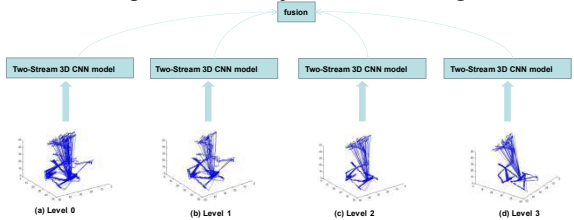

Fig. 3: Multi-Temporal structure for action *"hand waving"*

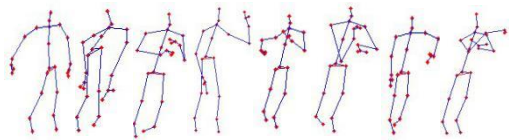

Fig. 4: Snaps from the NTU RGB+D dataset

*2) SmartHome Dataset:* SmartHome dataset [16] is collected by our lab, which contains six types of actions: "box", "high wave", "horizontal wave", "curl", "circle", "hand up". Each action is performed 6 times (three times for each hand) by 9 subjects in 5 situations: "sit", "stand", "with a pillow", "with a laptop", "with a person", resulting in 1620 depth sequences. Skeleton joints in SmartHome dataset contain much noises, due to the effect of occlusions and the unconstrained poses of action performers. For evaluation, we use subjects #1, 3, 5, 7, 9 for training and subjects #2, 4, 6, 8 for testing.

*B. Implementation Details*

We apply a normalization step on the joint coordinates by translating them to a body centered coordinate system with the "middle of the hip" joint as the origin. For the mapped 3D coordinate space $D$, we set the width height to be 50.

The network weights are learned using the mini-batch stochastic gradient descent with learning rate set to 0.0005, momentum value set to 0.9 and weight decay set to 1e-6. We use minibatches of size 32 and dropout with a probability of 0.3. The implementation is based on Keras [24] with one NVIDIA GeForce GTX 1080 card and 8G RAM. In addition, we sample randomly 10% of the initial training set as a validation set, which is used for hyper-parameter optimization.

## C. 3D CNN Architecture Evaluation

The spatio-temporal data augmentation method was used to solve the overfitting problem [22]. In our experiments, we adopt adding dropout layer (with p = 0.3) to receive convergence curves. Fig. 5 shows the convergence curves on the NTU RGB+D dataset for spatial stream and temporal stream, where the error rate tends to converge when the training epoch grows to 250. Since dropout can randomly drop units (along with their connections) from the neural network during training, it is able to prevent units from co-adapting too much.

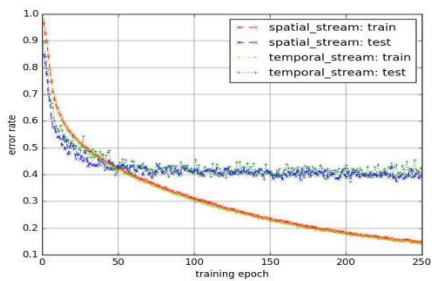

Fig. 5: Convergence curve on the NTU RGB+D dataset [17] for crossview evaluation

TABLE I: Evaluation of two-stream 3D CNN model

| Method | Dataset | | |
|---|---|---|---|
| | SmartHome (Cross Subject)(%) | NTU RGB+D (Cross Subject)(%) | NTU RGB+D (Cross View)(%) |
| Spatial Stream | 78.61 | 56.06 | 62.41 |
| Temporal Stream | 71.32 | 56.22 | 61.97 |
| Two-Stream | 79.38 | 62.13 | 67.87 |

TABLE II: Evaluation of multi-temporal architecture on NTU RGB+D dataset

| Method | Cross Subject(%) | Cross View(%) |
|---|---|---|
| Level 0 | 62.13 | 67.87 |
| Level 1 | 52.30 | 56.53 |
| Level 2 | 53.42 | 58.49 |
| Level 3 | 52.87 | 57.68 |
| Level 0+1+2+3 | 66.85 | 72.58 |

## D. Two-Stream 3D CNN Model Evaluation

Table I evaluates two-stream 3D CNN model method. By fusing the spatial stream and temporal stream, it has no obvious effect on SmartHome dataset for cross subject evaluation. Because SmartHome dataset does not contain action pairs that has opposite temporal order. On the contrary, two-stream 3D CNN respectively achieves 5.46% and 5.90% higher than individual stream on NTU RGB+D dataset for cross-view evaluation. These improvements verify the two-stream can mutually reinforce. Furthermore, some representative actions pairs like "sitting down" and "standing up". Its classification accuracy improved obviously. For spatial stream, the probability of classifying "sitting down" to "standing up" is 0.25. While for two-stream, the probability drops to 0.02. It can be seen that the error rate of mutual recognition has a reduction relatively.

## E. Multi-Temporal Structure Evaluation

The level $l$ of our two-stream 3D CNN model is considered to have notable impact on the performance. Table II show the recognition accuracies with different values of $l$ from 0 to 3. It also can be observed that our method achieves the best performance on the NTU RGB+D dataset when fusing all levels. In addition, four levels are used here, taking both performance and time cost into consideration.

Compared to other methods(see Table III) on the SmartHome dataset, the proposed two-stream 3D CNN model achieves the best performance, with the accuracy of 79.38%, which is better than Synthesized+Pre-trained [16]. Compared to ConvNets [25] and JTM [11], the improvements are 8.27% and 12.16% respectively. These improvements verify that our method can work well against noisy data.

TABLE III: Comparisons on the SmartHome dataset

| Methods | Year | Cross Subject(%) |
|---|---|---|
| ConvNets [25] | 2015 | 67.22 |
| JTM [11] | 2016 | 71.11 |
| SM+MM [26] | 2017 | 77.92 |
| Synthesized+Pre-trained [16] | 2017 | 78.61 |
| Two-stream 3D CNN(ours) | 2017 | 79.38 |

TABLE IV: Comparisons on the NTU RGB+D dataset

| Methods | Year | Cross Subject(%) | Cross View(%) |
|---|---|---|---|
| HOG2 [27] | 2013 | 32.24 | 22.27 |
| Lie Group [28] | 2014 | 50.08 | 52.76 |
| Skeletal Quads [29] | 2014 | 38.62 | 41.36 |
| FTP Dynamic Skeletons [30] | 2015 | 60.23 | 65.22 |
| HBRNN-L [10] | 2015 | 59.07 | 63.97 |
| Deep RNN [17] | 2016 | 56.29 | 56.29 |
| Deep LSTM [17] | 2016 | 60.69 | 67.29 |
| Part-aware LSTM [17] | 2016 | 62.93 | 70.27 |
| ST-LSTM [31] | 2016 | 61.70 | 75.50 |
| Multi-temporal 3D CNN(ours) | 2017 | 66.85 | 72.58 |

We compare the performance of our method with stateof-the-art methods on the NTU RGB+D dataset for crosssubject and cross-view evaluation and report the results in Table IV. Since this dataset provides rich samples for training deep models, e.g., HBRNN-L [10], achieved higher accuracy than most of handcrafted based methods. It verifies the effectiveness of the RNN-based methods. Besides, our method performs better than methods such as "Deep RNN" [17], "Deep LSTM" [17], Part-aware LSTM [17] for both crosssubject and cross-view protocols. And it also outperforms the ST-LSTM [31] for cross-subject evaluation and obtains competitive results for cross-view evaluation. Our method outperforms these methods mainly due to the following reasons. First, 3D convolutional neural network can sufficiently capture correlations , thereby learning spatial and temporal information simultaneously. Particularly, the form of spatial volume and temporal volume is useful for 3D CNN to represent information; Second, the network of two-stream enhances the spatialtemporal information and compensate for each other; Third, the multi-temporal structure learns multi-scale information including local patterns and global relationships.

## V. Conclusion and Future Work

This paper represents a two-stream 3D CNN model for action recognition based on skeleton sequences. The proposed spatial-temporal stream can learn more motion details of local and global by individual stream's mutual enhancement. Meanwhile, the improved 3D CNN architecture overcomes the overfitting problem and needs less training parameters than C3D. Experimental results show that our method outperforms most of state-of-the-art RNN-based approaches and verify the effectiveness of using 3D CNN learn the processed skeleton data, and that the multi-temporal version do increase the ability of 3D CNN model to capture multi-scale information. In the future, in order to be trained by 3D CNN more effectively, we will focus on different ways of encoding skeleton data.